# Scout Algorithm For Fast Substring Matching

Anand Natrajan[1], Mallige Anand

30 September, 2020

**Abstract:** Exact substring matching is a common task in many software applications. Despite the existence of several algorithms for finding whether or not a pattern string is present in a target string, the most common implementation is a naïve, brute force approach. Alternative approaches either do not provide enough of a benefit for the added complexity, or are impractical for modern character sets, e.g., Unicode. We present a new algorithm, Scout, that is straightforward, quick and appropriate for all applications. We also compare the performance characteristics of the Scout algorithm with several others.[2]

---

[1] Both authors conducted this work independent of any affiliation to any organisation.
[2] An abridged version of this paper has been submitted to Communications of the ACM for publication.

# 1. Introduction

Many software applications require an exact match of a pattern in a target string. Such matching is simpler than and distinct from matching regular expressions, contextual grammars, search engines, fuzzy matches and other related activities. In this paper, we will focus on exact matches, also called substring searches, alone. We will present an algorithm, dubbed "Scout". In most cases, this algorithm performs as well as or better than the best alternative substring search algorithms in existence. The algorithm is simple to implement, performs significantly better than the brute force algorithm most commonly employed, requires no preprocessing, has better memory usage characteristics than the best algorithms, and importantly, works for modern character sets, such as Unicode.

In our research, we also examined tests for comparing substring search algorithms. We were unable to find a standard set of benchmark tests. Consequently, it is difficult to reconcile the tests conducted by one survey with the tests conducted by another. Moreover, many of the tests focus on the number of character comparisons as the primary metric of performance. Our research indicates that memory lookups affect performance at least as much as comparisons. Finally, we consider some subjective measures for comparing various algorithms, for example, the suitability to other languages and non-ASCII character sets.

In order to compartmentalise several lines of inquiry, we will present this paper in three parts. In Part I, we will present the Scout algorithm, and contrast it briefly with related algorithms, such as brute force and Boyer-Moore. In Part II, we will present related work, as well as some augmentations that may improve upon them. Additionally, we will select exemplar algorithms for further comparison. In Part III, we will examine testbeds for performance comparisons. We will also examine factors that drive performance, and attempt to tease out the characteristics of the exemplars to explain their performance. Additionally, we will examine subjective factors that should influence our choice of algorithms for general-purpose substring searching. We will also present the results of performance tests. Finally, we will summarise this work. Past the references, we will present supplementary information such as code snippets, detailed tables, etc. in appendices to guide further research.

# Part I

In this part, we describe the Scout algorithm. We will walk through an example to describe the details of the algorithm. While we have conducted several tests to confirm that the algorithm works correctly, we will also provide a proof of correctness. We will end this part with a brief, subjective discussion about why the algorithm may perform better than alternatives, and why it may be a sound choice for general substring searching.

# 2. Description of Scout

The central idea in the Scout algorithm is to identify a "scout" character quickly. This character is dispatched to obtain information, in this case, an appropriate location in the target for a deeper match. A listing of the source code in Java is in [Appendix A: Java Source Code for Scout](). In order to describe this and subsequent algorithms, let us denote a pattern string $p$ as containing $m$ characters $p_0 p_1 p_2 \ldots p_{m-1}$, and a target string $t$ as containing $n$ characters $t_0 t_1 t_2 \ldots t_{n-1}$. The scout is always a character from the pattern string $p$, and we use it to find an appropriate location in the target string $t$.

Given $p$ and $t$, we begin the algorithm by comparing the first characters of each. On a match, we move on to the next characters of each, and so on, sequentially searching, much like the brute force approach. On a mismatch, Scout diverges from brute force. In brute force, we would re-initiate a sequential search comparison starting from the next character of $t$. In other words we would "slide" the pattern $p$ one character "to the right" along $t$. In contrast, in Scout, we designate the currently-mismatched pattern character as a "scout". We compare the scout with the character in *t* immediately past the mismatch. If those characters are mismatched, we advance along *t*, and compare the next character to the scout. We proceed in this fashion until the first time we encounter a character in *t* that matches the scout. If we find no such match, *p* is not contained in *t*. If we do, we slide *p* along *t* such that the last mismatched character in *p* (which was the scout's original position) lines up with the scout's current match in *t*. We then initiate a sequential search of *p* in *t* starting from this new position. An additional check, called a "twin" check can occasionally slide *p* along *t* further. We will return to this additional check later.

The Scout algorithm is somewhat intricate, so we walk through an example to understand how this algorithm works. Consider *t* and *p* as the target and pattern strings respectively (the indices in the top row are a visual aid alone):

```
        0           5              10             15             20
t =  a  a  a  c  b  a  b  a  a  a  b  c  b  a  a  b  c  a  a  b  a  c  a  b
p =  a  a  b  a
```

We begin in sequential mode, comparing $p_0$ = a with $t_0$ = a. Since those two characters match, we move on to $p_1$ = a and $t_1$ = a. Since they match as well, we move to $p_2$ = b and $t_2$ = a. Here, we encounter a mismatch. Consequently, we choose the current character in the pattern, $p_2$ = b as the scout (shown in bold font as a visual aid alone), and change into scout mode. In scout mode, we now compare $t_3$ = c with the scout character b. Since that comparison is a mismatch, we compare $t_4$ = b with the scout character b. Here, we encounter a match, so we slide the pattern *p* such that it aligns with a new position in the target *t*. The new alignment is shown visually below:

```
        0           5              10             15             20
t =  a  a  a  c  b  a  b  a  a  a  b  c  b  a  a  b  c  a  a  b  a  c  a  b
p =        a  **a**  b  a
```

Now, we revert to sequential search, and begin comparing from the start of *p*. We see that $p_0$ = a matches $t_2$ = a. Continuing to the next character, we see that $p_1$ = a fails to match $t_3$ = c. Accordingly, we choose the current pattern character, $p_1$ = a as the scout (again, in bold). The scout progresses a position at a time further along *t* to $t_5$ = a, where it finds a match. Again, we slide the pattern *p* such that it aligns with the new position in *t*.

```
        0           5              10             15             20
t =  a  a  a  c  b  a  b  a  a  a  b  c  b  a  a  b  c  a  a  b  a  c  a  b
p =              **a**  a  b  a
```

Reverting to sequential search again, we see that $p_0$ = a does not match $t_4$ = b. Therefore, we make $p_0$ = a the new scout (in bold) and slide the pattern one more position along the target before beginning sequential search:

```
        0           5              10             15             20
t =  a  a  a  c  b  a  b  a  a  a  b  c  b  a  a  b  c  a  a  b  a  c  a  b
p =                 a  **a**  b  a
```

Proceeding along, eventually in sequential search we discover $p_1$ = a does not match with $t_6$ = b so we enter scout mode with character a (in bold). In this case, scout mode yields a right shift of one position. Another foray into sequential search, another mismatch for $p_0$ = a, and another scout yields another right shift by one position.

```
        0           5              10             15             20
t =  a  a  a  c  b  a  b  a  a  a  b  c  b  a  a  b  c  a  a  b  a  c  a  b
p =                       a  a  **b**  a
```

The next scout would be $p_2$ = b (in bold) yielding a rightward shift of two characters. The next scout would be $p_3$ = a (not shown) for another rightward shift of two characters.

```
        0           5              10             15             20
t =  a  a  a  c  b  a  b  a  a  a  b  c  b  a  a  b  c  a  a  b  a  c  a  b
p =                                **a**  a  b  a
```

Yet another mismatch at $p_0$ = a (in bold) and another scout-assisted rightward slide by three positions leads to:

```
        0           5              10             15             20
t =  a  a  a  c  b  a  b  a  a  a  b  c  b  a  a  b  c  a  a  b  a  c  a  b
p =                                            a  a  b  **a**
```

The next scout is $p_3 =$ a (in bold) which encounters a mismatch with $t_{17} =$ c. The subsequent scout-assisted slide yields one character, leading to the following interesting alignment.

```
          0           5              10             15             20
t  =  a   a   a   c   b   a   b   a   a   a   a   b   c   b   a   a   b   c   a   a   b   a   c   a   b
p  =                                                              a       a   b       a
```

Per usual, sequential search informs us that $p_0 =$ a matches $t_{15} =$ a. However, for the first time in this example, we notice that $p_0 =$ a precedes and matches the scout character, i.e., the last a in the pattern (both in bold). As alluded to earlier, whenever we initiate a sequential search of $p$ in $t$ character by character, on a match, we additionally check if the current character in $p$ is identical to the scout and precedes it. If so, this character, which we call a "twin", can be used to slide $p$ further along until the twin aligns with the scout's position. Whether a slide occurred because of a twin or because of the scout alone, we initiate a sequential search of $p$ starting from its current alignment with $t$. During this sequential search, we may encounter another mismatch. This mismatched pattern character is the new scout, which continues the algorithm as before. Therefore, we have our first opportunity to invoke a twin-assisted slide, making $p_0 =$ a slide right to occupy the scout's position.

```
          0           5              10             15             20
t  =  a   a   a   c   b   a   b   a   a   a   a   b   c   b   a   a   b   c   a   a   b   a   c   a   b
p  =                                                                      a       a   b       a
```

This slide happens to align the pattern with its position the target, for a successful match. Note, the fact that the twin match happened between the first and last characters of the pattern is a coincidence, and not germane to the algorithm.

## 3. Proof of Correctness

LEMMA I: *Sequential search finds a match, or the first mismatched pattern character.*
This lemma is self-evident, so we accept it without formal proof. If the target is exhausted before the first mismatched pattern character is found, it is a non-match.

LEMMA II: *The scout character finds the first possible target location where a match could occur.*
This lemma is true by construction. The scout character is generated by a mismatch. The scout character moves along the target until it finds a match for itself in the target. Therefore, the scout's location represents the first possible match from its own starting location. If the target is exhausted before the first mismatched pattern character is found, it is a non-match.

LEMMA III: *Sequential search and scout alignment suffice to find a pattern match if it exists.*
Sequential search is clearly a necessary technique for every algorithm. By LEMMA I, if there is no match, sequential search identifies the first mismatched character. By LEMMA II, this character finds its first match in the target. A recursive application of this same process yields either a match or no match if the target is exhausted.

THEOREM I: *The twin character can be moved to the new position of the scout character.*
We prove the theorem by contradiction.
By LEMMA I, the sequential search finds the first mismatch from the current alignment of the pattern and the target. This mismatch is denominated as the scout.
Let $s_o$ be the old position of the scout, i.e., the position of the target character that did not match a pattern character per a sequential search.
Let $s_n$ be the new position of the scout, i.e., the position of the target character that first matched the scout character per a scout search.
By construction, $s_o < s_n$, i.e., for the scout, the new position succeeds the old position.

By LEMMA II, the scout character identifies the first possible target location thereon, thus moving from an old position to a new position.

Formally, there exists no $k$ such that $s_o < s_o+k < s_n$, where $t_{so+k} = t_{sn}$.

Let $w$ be the position of the target character that currently matches the twin character.

By definition, $w < s_o$, i.e., the twin character precedes the scout character.

By transitivity, $w < s_o < s_n$.

Assume an intermediate position $w+k$ in the target prior to the new scout position where the twin could match.

Therefore, $w < w+k < s_n$, where $t_{w+k} = t_{sn}$.

Colloquially, we say the twin character could move $k$ positions rightward. Therefore, even the scout character must move $k$ positions rightward.

Therefore, $s_o < s_o+k < s_n$, where $t_{so+k} = t_{sn}$.

However, this statement contradicts LEMMA II. If an earlier position existed, the scout would have found it. Therefore, either an earlier position does not exist, or $w+k = s_n$. In other words, the twin must move to the new position for the scout.  □

THEOREM II: *The Scout algorithm will find a pattern match if one exists.*

By LEMMA III, sequential search and scout alignment suffice to find a pattern match if one exists. During the sequential search, if a twin is never found, the algorithm will find a match if one exists. If a twin is ever found, per THEOREM I, it is always legitimate to align the twin to the new position for the scout and resume sequential search and scout alignment, which suffice. Therefore, in all cases, the algorithm will find a match if one exists.  □

## 4. Discussion of Performance

Strictly measured by character comparisons alone, Scout *does not* advance the field of substring matching. However, we expect better algorithms to execute faster, and by that metric, Scout indeed does clearly improve over most algorithms. Our explanation for this seeming paradox relies on hypotheses based on computer architecture.

The Scout algorithm attempts to skip over parts of the target by making as few comparisons as possible. All such attempts can outperform brute force simply by making fewer comparisons. Boyer-Moore makes far fewer comparisons, therefore outperforms brute force significantly. Scout makes roughly as many comparisons as brute force, yet outperforms it, to the extent that Scout and Boyer-Moore rival each other in performance. We hypothesise that Scout's performance gains come not from making fewer comparisons, but from two unusual factors: memory lookups and caching. As compared to brute force, Scout often requires half as many memory lookups. Boyer-Moore often requires even fewer memory lookups, e.g., a quarter to tenth as many as brute force. On most microprocessors, character comparisons are cheap but memory lookups are expensive, therefore performing fewer memory lookups entails significant speedup. Scout gains an additional edge in performance because the manner in which it accesses memory exploits temporal and spatial locality. Therefore, caching systems serve Scout better than they do Boyer-Moore. The net result is that Scout's performance, as measured by elapsed time, rivals the best algorithms in existence. We will examine performance in more detail in subsequent parts.

# Part II

In this part, we refer to related work in substring matching. It is not our intent to survey all available algorithms exhaustively. Instead, we pick several algorithms that may be considered best-of-breed so that we can compare them to Scout in later parts. In the course of this research, we were able to suggest some variants to some of these algorithms that improve the performance characteristics of those algorithms. Selecting those variants favours those algorithms in our subsequent performance comparisons. Although we have presented our comparisons against those variants alone here, our unpublished results include comparisons against the originals as well. None of the conclusions we draw here change because of our decision to select the best variants.

## 5. Related Work

The brute force, double-loop algorithm is the most common one used to find a substring. Since the late 1970s through the early 1990s, several alternative algorithms have been proposed, notably, Knuth-Morris-Pratt [KMP77], Karp-Rabin [KR87], Boyer-Moore [BM77] and Aho-Corasick [AC75]. Some of the best-performing algorithms known, i.e., the Horspool algorithm [Hor80] and the Sunday Quick Search algorithm [Sun90] are variants of Boyer-Moore. Several surveys of these and other classes of algorithms can be found in the literature [CL97]. Although the theoretical average- and worst-case performance of each of these algorithms beats brute force, these algorithms have not found their way into common usage. We believe that there are several criteria other than theoretical performance that hinder their adoption. We will examine those criteria in more detail in a subsequent section.

Most substring search algorithms fall into a few broad classes. One large class contains algorithms that identify properties of the pattern or target so that portions of the target can be skipped over. This class includes Boyer-Moore and several variants. Most of these algorithms require a preprocessing step. Most of them reserve memory proportional to the size of the pattern or the size of the alphabet used in the pattern and target. Brute force and Scout may be considered degenerate members of this class. Brute force skips characters by the minimum possible value, i.e., one. Brute force and Scout require no extra memory. A second class contains algorithms that compute some heuristic that could result in false positives, but not false negatives. On a positive match for the heuristic, these algorithms fall back to a sequential search for disambiguation. This class includes Karp-Rabin and some variants. A third class contains algorithms that use characteristics of the language, e.g., character frequencies, to determine how to navigate the target. This class includes Aho-Corasick and other variants. Yet other classes may exist. We look at related work more closely in the next part.

As mentioned earlier, most substring search algorithms either attempt to skip comparing portions of the target, or compute heuristics that tolerate false positives, or use language features to speed up the process of finding a pattern in a target. In our work, we do not undertake any objective comparisons against algorithms in the third class, namely, language-specific ones. Our goal has always been to find the best possible generic substring matching algorithm, not one specific to any domain.

The brute force algorithm is deployed widely. It is the String `indexOf` algorithm deployed in the Java language library for OpenJDK as of version 9.0.1. In other words, one of the most common libraries for one of the most popular languages for large-scale applications uses the brute force approach for any and all substring matching. The C implementation of `strstr` also uses brute force. The Python implementation of the `find` method uses Sunday Quick Search (although it claims to use Boyer-Moore), but makes two key accommodations in order to circumvent the problem with that algorithm, namely the need to store a bad character array whose size is dependent on the character set. Instead of storing the bad character skip value for all characters in the set, the skip value for only the last character is stored. Also, the Python implementation checks for presence of a target character in a pattern with a low-fidelity Bloom filter that can result in false positives, e.g., treating à the same as ā. These accommodations result in smaller shifts than the classical Boyer-Moore algorithm, but avoid some of its most problematic characteristics.

## 6. Rolling Algorithms

We offer two Rolling algorithms as variants to Karp-Rabin. The central idea in these algorithms is to perform a quick and cheap match first. The cheap match may result in a false positive, which must then be disambiguated from a true positive with a sequential search. We offer two Rolling algorithms: Rolling Sum and

Rolling XOR. In order to describe these algorithms, let us denote a pattern string $p$ as containing $m$ characters $p_0 p_1 p_2 \ldots p_{m-1}$, and a target string $t$ as containing $n$ characters $t_0 t_1 t_2 \ldots t_{n-1}$.

### 3.1 Rolling Sum

In the Rolling Sum approach, given $p$ and $t$, we compute a pattern signature for $p$, $\sum p_{0..m-1}$, as the **sum** of integer values for each of those characters. The integer value for a character may best be thought of as the Unicode value for that character. Turning to the target string, we pick a window of the first $m$ characters (assume $m \leq n$) and compute a target signature, $\sum t_{0..m-1}$. If the target signature does not equal the pattern signature, we move the window to the next $m$ characters, i.e., we compute a new target signature, $\sum t_{1..m}$. However, at this point, instead of recomputing the signature from scratch, we observe that:

$$\sum t_{1..m} = \sum t_{0..m-1} - t_0 + t_m.$$

In other words, the signature for the next window in the target can be computed by modifying the signature of the previous window, like in Karp-Rabin. We now compare the new target signature to the invariant pattern signature, sliding the target window along whenever the current target signature does not match the pattern signature.

Of course, it is possible that the current target signature matches the pattern signature even though the substring is not a match. Such a false positive can occur because of swapped character positions, or even different subsets of characters accidentally yielding the same signature. In such cases, the Rolling Sum algorithm falls back to a brute force approach, which incontrovertibly disambiguates the true and false positives. On a false positive, the algorithm continues sliding the target window. A true positive is a successful match. The process continues until a successful match is found or the target string has been exhausted.

We offer a pseudocode implementation of Rolling Sum below, with edge-case checks elided:

```
given target string t, pattern string p
let n = target.length, m = pattern.length
let patternSig = 0, targetSig = 0
for i from 0 to m - 1
    patternSig += integer value of pattern[i]
    targetSig += integer value of target[i]
for i from 0 to n - m
    if patternSig == targetSig
        compare pattern[0 .. m-1] to target[i .. i+m-1]
        if matched return success
    // if here, either signature didn't match or it did
    // but brute force failed.
    targetSig -= integer value of target[i]
    targetSig += integer value of target[i + m]
return failure
```

The four bolded lines represent the heart of the Rolling Sum algorithm. Those lines could raise three possible concerns. One, the arithmetic operations could be expensive. On most modern microprocessors, integer addition and subtraction are implemented with one or two processor cycles, i.e., in the fastest possible manner. In other words, the arithmetic operations are about as expensive as character comparisons or loop variable increments. Of course, when the signatures *do not match*, the algorithm obviates large numbers of character comparisons, thus providing practical speed-up. Two, the signatures could have too many false positives, devolving into a brute force check but with the overhead of computing the rolling sum. Certainly, it is possible to create pathological test cases that demonstrate that concern (e.g., imagine searching for the pattern "`ac`" in a target "`bbbbbbbb`"), but practically, we expect false positives to be relatively rare. The signature can be made more discriminatory if we used multiplication/division instead of addition/subtraction, but doing so substitutes cheap +/- operations with expensive ×/÷ operations. Three, the signatures could overflow the data type used to compute them. However, as long as the overflow does not cause an exit, it does not matter. An overflow would result in a negative sum, which is perfectly valid as a signature.

### 3.2 Rolling XOR

Some of the concerns with the Rolling Sum approach can be alleviated with a Rolling XOR approach. The Rolling Sum works because the function used to compute the signature is reversible, commutative and cumulative. It is reversible because addition has a dual, namely, subtraction. It is commutative, therefore integer values for characters can be added and subtracted in any order to retrieve sub-signatures. Finally, because the function is cumulative, it is a compressed representation of the pattern, albeit with a propensity to false positives. As we alluded earlier, but for the cost, multiplication could also be chosen for the signature, as it too is reversible, commutative and cumulative.

Another function that is reversible, commutative and cumulative is the bitwise XOR function. Accordingly, in the Rolling XOR approach, given $p$ and $t$, we compute a pattern signature for $p$, $\veebar p_{0..m-1}$, as the bitwise **XOR** of integer values (Unicode) for each of those characters. Turning to the target string, we pick a window of the first $m$ characters (assume $m \leq n$) and compute a target signature, $\veebar t_{0..m-1}$. If the target signature does not equal the pattern signature, we move the window to the next $m$ characters, i.e., we compute a new target signature, $\veebar t_{1..m}$. However, at this point, instead of recomputing the signature from scratch, we observe that:

$$\veebar t_{1..m} = \veebar t_{0..m-1} \veebar t_0 \veebar t_m.$$

In other words, as before, the signature for the next window in the target can be computed by modifying the signature of the previous window. The Rolling XOR algorithm's performance characteristics are also $O(m + n)$.

The pseudocode implementation of Rolling XOR is near-identical to the Rolling Sum above. The only changes are to the bolded lines. Those changes are shown below:

```
given target string t, pattern string p
...
    patternSig ^= integer value of pattern[i]
    targetSig ^= integer value of target[i]
...
    targetSig ^= integer value of target[i]
    targetSig ^= integer value of target[i + m]
return failure
```

Revisiting the concerns with the Rolling Sum algorithm, some benefits of the Rolling XOR algorithm are apparent. One, the signature operations are cheap. On most modern microprocessors, bitwise operations are implemented with one processor cycle, i.e., in the fastest possible manner. Two, the signatures could still have too many false positives. It is unclear which of Sum or XOR raises more false positives; for example, the "`ac`" in "`bbbbbbbb`" case above is a false positive for Sum, but not for XOR. However, conceptually an XOR operation can have one fewer result bit as compared to Sum (the carry bit), therefore carries less information than Sum. Three, overflows are not an issue for XOR, because there is no carry bit.

## 7. Scout Variants

In the course of crafting the Scout algorithm, we explored several variants. These variants are attractive because they either simplify the algorithm, or avoid repeated recalculations of some items, or combine features of the Scout algorithm with an alternative algorithm.

## 7.1 Scout Simple

The Scout Simple variant of the Scout algorithm simplifies the logic within the Scout algorithm, but at a cost to performance. Given $p$ and $t$, we begin the algorithm by comparing the first characters of each. On a match, we move on to the next characters of each, and so on much like the brute force approach. However, on a mismatch, In Scout Simple, we designate the last character of $p$ as the scout. We march the scout along the target exactly as in the Scout algorithm. If we find no match, $p$ is not contained in $t$. If we do, we slide $p$ along the target such that the last character in $p$ (which is at the scout's original position) lines up with the match in $t$. We then initiate a full search of $p$ in $t$ starting from this new position. However, we never initiate the twin matching portion of the Scout algorithm. Once again, if a mismatch occurs we designate the last character in $p$ as the scout and repeat the process.

The algorithm is fairly simple, and captures much of the performance gains of the preceding Scout algorithm, but not all of it. In terms of comparisons, the Scout Simple algorithm missed out on some opportunities to find twin characters because the scout is always the last pattern character. Memory lookups are virtually identical between both algorithms. We hypothesise that the loss in performance is because of poorer spatial locality. In Scout, when a mismatch occurs on a sequential search, the scout search takes over from the very next character in the target. In contrast, in Scout Simple, when the sequential search fails, the scout begins at a potentially faraway point in the target.

## 7.2 Scout Twin

In the Scout Twin variant of the Scout algorithm, we observe that if any given character in the pattern has a twin, it is always the first occurrence of that character in the pattern. Accordingly, Scout Twin introduces a preprocessing step to compute and store the positions of those twins once. The algorithm proceeds near-identically as the Scout algorithm, beginning with sequential search, encountering a mismatch, selecting a scout and marching it forward. In Scout Twin, we straightaway identify whether this scout has a twin or not, and where it is located in the pattern. If a twin exists, the pattern can be slid further.

Alluring as the Scout Twin is, it performs worse than Scout. First, it requires a preprocessing step to locate the twins. This step has $O(m^2)$ complexity. Second, the twins have to be stored in memory in an array whose size is $O(m)$. Third, accessing this array during the search introduces a memory lookup with poor spatial or temporal locality.

## 7.3 Scout Variant

The Scout Variant variant of the Scout algorithm is a minor adjustment to the Scout algorithm that offers no theoretical advantage, and ends up being poorer in practice. In Scout, we observe that after finding a scout and sliding the pattern to align with the scout, we embark on a modified sequential search. We check each pattern character against a corresponding target character, and if they match, we check the pattern character against the scout to see if it is a twin. The Scout Variant reverses the order of those two checks. In this variant, we first check if a pattern character is a twin, and only then check against the corresponding target character.

The Scout Variant has roughly the same number of memory lookups and comparisons as Scout. However, its performance is slightly worse, likely because of artifacts of the underlying microcomputer architecture.

## 7.4 Scout Sunday

The Scout algorithm resembles the Boyer-Moore class of algorithms in the sense that all of them exploit a mismatch between a pattern character and a target character as an opportunity to slide the pattern along the target. The Scout algorithm identifies a scout in the pattern string to determine how much to slide. The Boyer-Moore algorithms use the mismatched target character, specifically, its last occurrence in the pattern to determine how much to slide. Of these, the Sunday Quick Search algorithm does not depend on the order in which the pattern is searched in the target, i.e., first-to-last or last-to-first order does not matter. This algorithm also is the Boyer-Moore variant that moves the pattern along most aggressively, and is therefore the fastest under most circumstances.

Our proposed Scout Sunday algorithm combines either Scout, Scout Variant or Scout Simple with Sunday Quick Search. All of these algorithms begin with a sequential search. On a mismatch, our proposed Scout Sunday algorithm would perform the target-based bad character slide suggested by the Sunday Quick Search algorithm, then attempt an additional slide based on the scout character. We believe this amalgamated algorithm could have attractive theoretical performance. However, we have not analysed it in detail. First, we have not attempted a proof of correctness for this approach. Second, we have not encoded all of the combinations mentioned above in order to conduct empirical observations. Finally, even if we undertook these tasks, Scout Sunday would suffer from some of the challenges faced by all of the Boyer-Moore class of algorithms.

## 8. Exemplar Choices

When comparing and contrasting substring search algorithms, we face an abundance of choices. At the time of this writing, we are aware of over forty algorithms for substring search, including brute force and all of the algorithms and variants we have listed here. Exhaustive comparisons across all of these based on performance, memory, applicability, simplicity, etc. would be a welcome body of work. In the next part, we make a modest contribution towards that body of work without taking it on in full. Our more detailed comparisons will use the following algorithms alone: Brute force, Rolling XOR, Sunday Quick Search and Scout. While we acknowledge the risk involved in limiting our focus to a few algorithms alone, we believe our choices are reasonable because they permit us to perform a deep analysis along our lines of inquiry.

We chose brute force simply because it is a baseline as well as an incumbent algorithm. We chose Rolling XOR to represent the heuristic class of algorithms because it performs far better than its peers in that class, yet does not incur some of their expensive operations. To put matters in perspective, Karp-Rabin, which could lay claim to being an exemplar of this class almost always performs worse than brute force, whereas Rolling XOR beats brute force. The Boyer-Moore class has the largest number of variants, but it is generally accepted that Sunday Quick Search is the fastest among them. We did not choose language-specific algorithms because our goal has always been to identify domain-agnostic algorithms. Finally, we chose Scout from among our many variants because it performed the best.

# Part III

When comparing substring search algorithms, most authors focus on performance, understandably so. Performance often translates to number of character comparisons, whether best, worst or average case. Of course, other metrics matter as well, for example memory usage. However, character comparisons dominate most discussions around performance because character comparisons count towards order complexity, and order complexity remains the gold standard in computer science for evaluating the performance of algorithms. We argue for a broader approach for evaluating substring searching algorithms, without diluting the focus on performance.

Our performance numbers indicate that Scout should be a strong contender for a general-purpose substring searching algorithm. As a matter of practical performance, we will show that Scout is one of the fastest algorithms available. While it may score poorly on the count of character comparisons, it scores well on memory lookups. Although traditional measures of performance are based on character comparisons, we believe memory lookups should be given more prominence because the former are far less expensive than the latter. Therefore, an algorithm that reduces memory lookups, be it at the expense of more character comparisons is likely to be faster, as shown by Scout.

In our experiments, we have endeavoured to offer every advantage possible to alternative algorithms. We have faithfully transcribed the implementations available publicly. We made variable names and formatting more readable, which does not affect performance. Most importantly, for algorithms that required memory proportional to the alphabet size, we assumed the alphabet size to be 256 characters. In other words, we permitted these algorithms to function for the ASCII character set alone, although Scout works as-is for Unicode. When we refactored the alternative algorithms faithfully to work for Unicode, they either failed outright or resulted in absurdly high wall-clock times.

Several subjective criteria may influence the choice of a general-purpose substring search algorithm over and above practical performance, preprocessing and memory usage. Below, we discuss a few other criteria and argue for attention towards them. In [Appendix C: Tabular Comparison of Various Algorithms](Appendix C: Tabular Comparison of Various Algorithms), we will list out all of the algorithms we have surveyed extensively, i.e., implemented and tested against, and present our assessment of those algorithms as well as several others against the criteria below.

- *Character set*. Some algorithms, notably Karp-Rabin and Boyer-Moore, assume an ASCII character set consisting of 256 characters, i.e., characters representable within one byte. Modern Unicode character sets can take 1-4 bytes. These algorithms were formulated when Unicode was not prevalent. With Unicode, they either fail outright, or when upgraded, consume prohibitive amounts of memory.
- *Language specificity*. Some algorithms, notably Aho-Corasick and Sunday Maximal Shift rely on letter frequencies in languages. Obviously, letter frequencies vary by language, making the algorithms harder to generalise. Moreover, letter frequencies in specific subdomains will likely differ, e.g., in genome matching.
- *Simplicity*. The brute force approach is simple to implement, requiring less than 20 lines of formatted code, including boiler-plate. The other approaches, including ours, can be 2-4 times larger. Of course, lines of code is a trivial concern, but conceptual complexity can complicate provability, thus hindering broader acceptance.

## 9. Practical Performance

The point of evaluating performance is to assess whether one algorithm is speedier than another, especially at scale. Turning our attention to time taken as a metric of performance, we recognise that while character comparisons certainly contribute towards time taken, they are not as dominant as other factors. Several algorithms undertake preprocessing steps in order to speed up subsequent searching. The work done in those preprocessing steps technically does not count towards the number of character comparisons. However, general-purpose substring searching is likely to have low reuse, i.e., it is likely that every time the algorithm is invoked, the pattern and target may change. Therefore, algorithms that preprocess either the pattern or target or both have to undertake their preprocessing steps for every single invocation. Preprocessing consumes time, whether or not the character comparison counts improve.

Yet other algorithms involve expensive operations, e.g., modulo arithmetic. On most microcomputers, these operations are far more expensive than character counts. Certainly, some microcomputers may come equipped with separate processing units to speed up arithmetic calculations, but such processing is still likely to be more expensive than character comparisons.

Memory lookups affect performance. Given how cheap character comparisons are on most modern microcomputers, memory lookups can take several more processor cycles than character comparisons. Therefore, reducing the number of memory lookups can reduce time taken to execute an algorithm. Memory lookups play another role in performance when spatial and temporal locality are taken into account. Techniques such as paging and multi-level caching have been developed to reduce the cost of memory lookups. While we do not expect to craft algorithms to account for specific paging or caching techniques on specific machines, an algorithm that naturally exploits basic paging and caching will perform better than an equivalent algorithm that does not.

We have chosen to normalise all of these factors by comparing wall-clock time. The question of how to account for all of these and other factors is challenging. Focusing on character comparisons alone can seem myopic especially if a low character comparison count comes at the expense of too many memory lookups or costly arithmetic operations. Loading up most of the work in a preprocessing step so as to make the search cheap is counter-productive. Deep textual analysis may reduce character comparisons dramatically, but may not lend themselves to practical solutions. By focusing on wall-clock time, we force a normalisation of all of these disparate factors into a common currency that is comparable and practical.

Wall-clock time as a common currency is not flawless. It will be readily apparent that wall-clock times for the same tests run on different machines, operating systems, compilers, loads, etc. will differ. Therefore, only relative wall-clock times lead to meaningful comparisons. More particularly, the wall-clock times have to be compared for the same tests on the same machines, under the same load conditions. Another issue with wall-clock times is that underlying performance trends and patterns are not readily apparent unless several data points are plotted and examined. For example, a pen-and-paper analysis of an algorithm may suggest its performance grows linearly, but a wall-clock time plot may reveal large start-up costs that are not amortised well enough to make the linear growth apparent until extremely large data sets are chosen. Alternatively, another seemingly linear-growth algorithm may be revealed to encounter page thrashing, which causes performance to deteriorate for large data sets.

## 10. Standard Testbeds

We were unable to find a standard set of benchmark tests for assessing substring search performance. Therefore, we crafted a synthetic suite of tests, written in Java. Our testbeds were crafted to test several hypotheses:.

1. Obviously, our algorithms as well as the alternatives should function correctly. We measured correctness somewhat more stringently than necessary, by checking for the exact position at which a pattern was found in the target, rather than merely a true/false check. We also checked for Unicode patterns and targets.
2. We varied the depth at which the pattern was found in a fixed target. In other words, we crafted a synthetic target of a fixed size (100 characters), and inserted a fixed-size pattern (5 characters) that was guaranteed to exist in the target. We varied whether the pattern was found at the start (0%), the end (100%), the middle (50%) as well as several other intermediate positions. Specifically, our pattern string was "`aabca`", and our target string was "`xx...xaabcaxx...x`", where the prefix of `x` characters was of length $p$, and the suffix was of length $q$. Therefore pattern length $m = 5$, and target length $n = p+q+m = 105$. If $p = 0$, the pattern depth is 0%, but if $q = 0$, the pattern depth is 100%. Using this testbed, we tried to ascertain how algorithmic performance varied from best case (0%) through worst-case (100%).
3. We varied the length of the target string. We crafted increasingly long target strings (from 0 through 10,000 characters), and appended a fixed pattern string (5 characters) at the very end to simulate worst-case behaviour. Specifically, our pattern string was "`aabca`", and our target string was "`xx...xaabca`", where the prefix of `x` characters was of length $p$. Therefore pattern length $m = 5$, and target length $n = p+m$. We varied $p$ in order to craft longer target strings. We expected all algorithms to grow linearly in wall-clock time as length increased. However, we wanted to observe the slopes of those linear curves for various algorithms.
4. We simulated real-life scenarios by choosing a well-known literary passage (Hamlet's famous soliloquy), and searching for substrings deeper and deeper into this target, again starting from 0% through 100%. This test not only simulated real-life human behaviour, but also tested whether all of the algorithms processed punctuation, spaces and mixed cases correctly. By concatenating all of the lines in the soliloquy, we were able to craft a target string of modest-to-large size (1500 characters). The exact target string, as well as our test cases are shown in Appendix B: Real-life Testbed. We readily concede the cultural monotony in the choice of this text. While it does not include characters from other languages or even accents, by choosing a well-known piece of English text, we wished to give language-specific algorithms whatever advantage they could claim.

For all of the tests, in our Java implementations, we measured four metrics. One, wall-clock time, with timers surrounding repeated method invocations within loops. Two, comparisons, by instrumenting the code with counters. Three, memory accesses, for array characters in the pattern and target string. In short, we counted one memory access for every pair of [ and ] brackets in the code, but not for accesses to local or global variables. Four, heavy arithmetic operations, i.e., multiplication, division and modulo operators (and exponentiation had we encountered it). We did not count bitwise operators as well as addition and subtraction. Not only are those operators inexpensive, but had we accounted for those, we would have to add up array offsets, loop variables, negations, etc. The instrumentation for comparisons, memory accesses and heavy arithmetic slows down the raw performance of the algorithms, which is another reason to ignore *absolute* wall-clock times but pay attention to *relative* times.

We did *not* measure memory consumption in our tests, primarily so we could focus on practical performance. Of course, memory consumption matters, and several alternative algorithms store large amounts of global or per-pattern or per-target state prior to the actual search. We have penalised all preprocessing by counting them towards the wall-clock time taken to execute the search. Doing so exempts algorithms that require global preprocessing, e.g., constructing a one-time dictionary, and also ignores the memory consumption of these algorithms. Other surveys have examined the memory consumption characteristics of alternative algorithms. We refer interested readers to those surveys. Briefly, we will note that of our exemplars, viz. Brute force, Rolling XOR, Sunday Quick Search and Scout, only Sunday Quick Search requires any significant memory. This search algorithm, like most members of the Boyer-Moore class, stores a bad-character integer map whose size is proportional to the size of the character set.

## 11. Methodology and Results

Our tests ran on a MacBook Pro with a 2.9 GHz Intel Core i9, with 32GB RAM, running macOS v10.15.7 (Catalina). The Java version was 9.0.1 (build 9.0.1+11). Where applicable, our C compiler was Gnu gcc (clang-1200.0.32.2), and our Perl interpreter was version 5 (v5.30.1).

Each algorithm compared was coded fairly, using open-source Java implementations. For the baseline brute force algorithm, we reimplemented the algorithm rather than relying on the underlying Java implementation in order to eliminate low-level compiler optimisations. Every implementation was checked for correctness against several cases. The algorithms were "touched" lightly to make variable names consistent and readable, and formatting consistent. One conundrum we faced was with the fact that several of the algorithms, in particular Karp-Rabin and all of the Boyer-Moore variants, tolerated only ASCII characters. In such form, their performance was good, but if re-coded to permit Unicode character sets, the algorithms were entirely unworkable. For example, the Boyer-Moore algorithms would require an integer array containing $2^{32}$ entries *for every test*. Aside from the fact that such an array would use up half of the 32GB of memory on our test machine, such an algorithm is impractical for general-purpose use. In the end, we left the algorithms as-is in the spirit of offering every advantage to alternative algorithms.

In most of our presentation below, we focus on the exemplar algorithms alone: Brute force, Rolling XOR, Sunday Quick Search and Scout. Doing so permits us to focus on the salient points of the comparison. We have unpublished results comparing several other algorithms as well; those results support the conclusions here. Choosing a few exemplars also permitted us to translate them to C and Perl so as to eliminate any peculiarities of Java, our preferred implementation language. Our results in these other languages again support our general conclusions, although the specific numbers between Java, C and Perl implementations vary because of compilation techniques.

Before we dive deeper into performance here, we will make some brief notes about several algorithms, with respect to their expected performance.

The Boyer-Moore class of algorithms, including Sunday Quick Search, are justly considered as some of the best at minimising the number of comparisons. Our algorithm, Scout, fares as badly as brute force in this regard. However, the Karp-Rabin class of heuristic algorithms, including Rolling XOR, easily surpasses all of the Boyer-Moore class of algorithms. Depending on the pattern and target, the Rolling XOR can end up with a comparison count equal to as little as the length of the pattern if present, or zero if absent.

The Karp-Rabin algorithm is the only one of the ones we surveyed exhaustively that incurs heavy arithmetic. Recall that Rolling Sum and Rolling XOR perform addition, subtraction and bitwise operations alone. The performance of these algorithms reflects those choices.

The algorithms based on constructing finite automata performed so poorly at load that we abandoned them after checking them for correctness. Running large tests with them would have produced absurdly large wall-clock times that would swamp out any meaningful comparison with anything else. Accordingly, for these algorithms, we do not attempt to measure or discuss auxiliary metrics such as comparisons or memory accesses.

Several algorithms, including the Boyer-Moore class, perform a preprocessing step, which we have chosen to include within the timing results. The time spent on preprocessing gets amortised over the time spent during the search. Put another way, if the target string is large, the time spent on preprocessing becomes negligible. However, for short target strings, the startup cost overhangs on the performance of these algorithms.

Our performance results are shown in the charts below. Detailed tabular data is in [Appendix D: Performance Tables and Graphs](). If plotting wall-clock time, the Y-axes are in nanoseconds. For each data point, we averaged the result of one million runs to smooth out any spurious results. Again, the absolute numbers are irrelevant; only the relative performance of the algorithms matters.

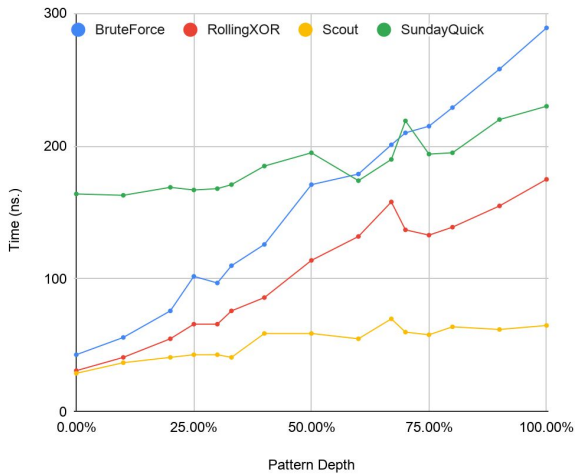
**Fig. 1:** Wall-clock times (ns.) for different pattern depths in a given target string length (100 characters).

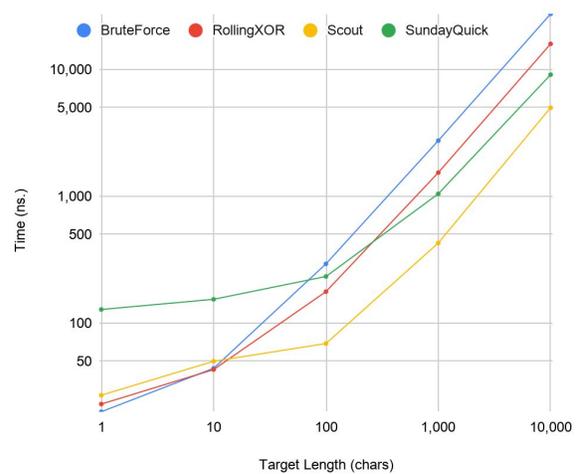
**Fig. 2:** Wall-clock times (ns.) for different target string lengths with a given pattern depth (100%). Log scales.

Scout clearly outperforms all of the other exemplars at most data points in terms of wall-clock time. All of the algorithms display roughly linear performance, but the slopes and intercepts are rather different. For a given target string, as the pattern is situated deeper and deeper (Fig. 1), we observe that Scout and Sunday Quick Search have gentler slopes, i.e., the time taken increases slowly as the pattern is found deeper in the target string. However, Sunday Quick Search has a large y-intercept, because of its high preprocessing cost. If the target string is increased in size, with the pattern situated at 100% (Fig. 2), all of the algorithms display a linear increase in time. The high preprocessing overhead penalises Sunday Quick Search for small target string lengths, but as the strings elongate, the benefits of that processing become apparent.

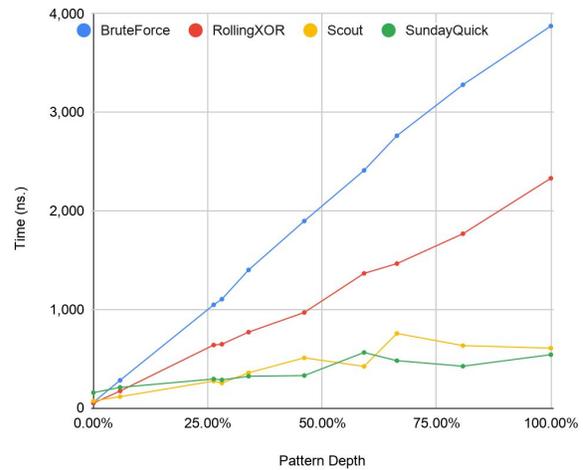

**Fig. 3:** Wall-clock times (ns.) for different pattern depths in Hamlet's soliloquy (1500 characters).

Karp-Rabin (not shown here), unfortunately pays too high a price for its expensive hash-coding algorithm, never approaching even brute force. However, Rolling XOR, which is in the same class as Karp-Rabin, holds its own against brute force, vindicating our faith in signatures whose quick computational speed outweighs their greater propensity towards false positives.

Extrapolating the trajectories of the Scout and Sunday Quick Search plots in Fig. 2 suggests that Sunday Quick Search will become the fastest algorithm if we consider target strings of length 100,000 characters or greater. To give a sense of the size of the target strings where Sunday Quick Search shines, note that a typical Shakespearean play, if rendered as a *single string*, would run from 150,000 to 500,000 characters. Our Hamlet testbed of course takes merely one soliloquy's worth of text. The performance characteristics of the various exemplars for this testbed are shown in Fig. 3. Scout and Sunday Quick Search trade top spots as the desired pattern is found deeper in the text. We note again that in these comparisons, we focus on wall-clock time alone, including preprocessing time, but excluding considerations such as memory consumption and ability to process Unicode text.

Our general results hold even if the implementation language is changed from Java to Perl or C even though specifics such as absolute time or the crossover points change. Overall, Scout continues to be the fastest algorithm of the exemplars. As we alluded previously, we have unpublished results where we perform similar rigorous comparisons of more algorithms, not just the exemplars. We present a sample of those results in tabular form in [Appendix D: Performance Tables and Graphs](#).

We now turn our attention to key performance factors. Throughout this paper, we have argued that character comparisons alone do not suffice for judging the performance of substring search algorithms. In the next few charts, we show the results of instrumenting our Java code to count character comparisons as well as memory lookups. (We also counted heavy arithmetic operations, but since those affect Karp-Rabin alone, we do not present them here.) We will consider the "real-life" Hamlet testbed for our discussion below.

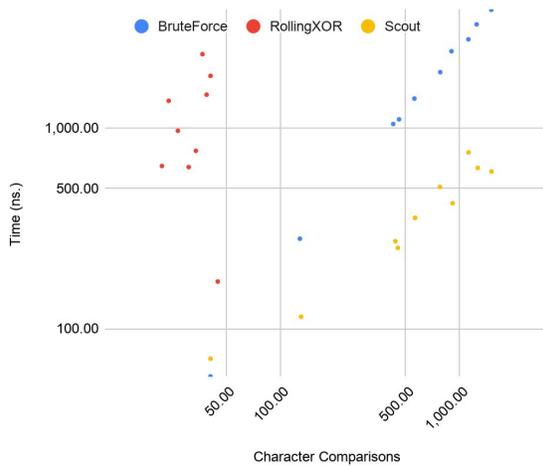 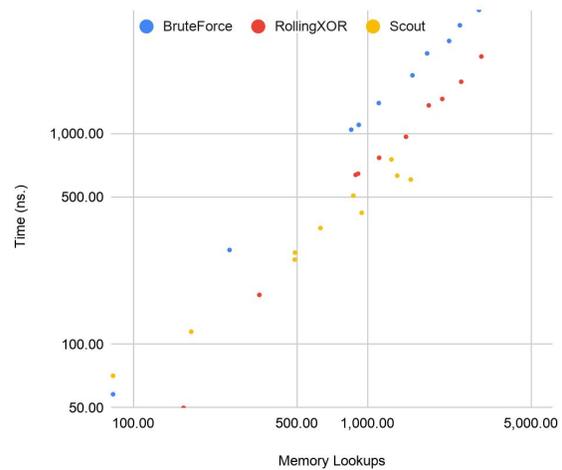

**Fig. 4:** Wall-clock times (ns.) at different counts of character comparisons for exemplar algorithms.

**Fig. 5:** Wall-clock times (ns.) at different counts of memory lookups for exemplar algorithms.

The scatterplot in Fig. 4 shows that wall-clock time is loosely linearly correlated with the number of character comparisons. We make two observations from this plot. One, when considering a particular algorithm alone, sometimes the correlation is tight (e.g., for Brute force), and sometimes loose (e.g., for Rolling XOR). Two, when considering all of the algorithms put together, the correlation is extremely weak. In contrast, the scatterplot in Fig. 5 shows a much tighter correlation in both cases between wall-clock time and memory lookups. Of course, character comparisons are correlated to memory lookups; the point of accessing pattern and target characters is to facilitate a comparison. However, whether considering a specific algorithm, or looking at all algorithms, memory lookups is a much better predictor of wall-clock time performance than character comparison counts.

Another key performance factor is spatial and temporal locality. An algorithm that accesses the same characters in rapid succession, or accesses nearby characters quickly, lends itself to better caching performance. We did not measure the effect of caching on any of the algorithms. However, an inspection of the code for the algorithms reveals that Scout shows good spatial and temporal locality. In contrast, Sunday Quick Search, despite performing fewer memory lookups, shows poor locality. In particular, as with every Boyer-Moore variant, it depends on the bad character array, which is simply another array of integers different from the pattern and the target, and which is indexed and accessed by the ASCII value of the currently-mismatched target character. The access pattern for the bad character array is effectively random, leading to poor locality. Although we were not able to quantify the effects of caching, we encourage further research into how caching can improve the performance of substring matching.

## 12. Conclusions

Searching for a substring is so routine a task that most languages provide in-built libraries that software developers can reuse. Many of these libraries implement a brute force algorithm to search for a substring despite there being dozens of alternative algorithms. We submit that these library methods should be rewritten to implement better algorithms. We provide an algorithm, Scout, whose performance as measured by wall-clock time ranks among the fastest algorithms. Additionally, Scout requires no preprocessing and has low, constant memory usage. It is language-agnostic and works for any character set.

We crafted testbeds so that we could compare Scout against the best alternatives. Along the way, we offered up some variants of existing algorithms, including Scout. We picked brute force and the best variants as exemplars for comparison. When performing our comparisons, we endeavoured to give every advantage to each exemplar, so as to make the performance comparisons salient.

Scout runs faster than the exemplars in most cases. Since we picked exemplars to favour different classes of algorithms, we are confident that Scout performs better than most alternatives. In unpublished results, we have verified this claim against additional alternatives.

In our research, we discovered that the usual metric of performance for substring matches, i.e., character comparisons, is a weak predictor of wall-clock time. Our results show that memory lookups predict performance better. We showed that merely counting memory accesses correlated better with observed wall-clock time. We speculate that caching effects because of spatial and temporal locality explain the remaining memory-related performance.

We encourage popular implementations of substring libraries to use our Scout algorithm.

# Appendix A: Java Source Code for Scout

We present a Java implementation of the Scout algorithm. The listing below focuses on the algorithm. It does not present surrounding code like class definitions, test cases and a main method, which can be furnished easily by any Java developer. Also, while we discuss a `contains` method in most of this paper, the listing below shows an `indexOf` method. In our code, as in most implementations, the `contains` method is simply a trivial check of the integer returned by the `indexOf` method. The integer returned corresponds to the first position of the pattern in the target or -1 if the pattern is absent. Finally, cursory inspection reveals that we chose to implement the algorithm using character arrays, rather than the built-in String class in Java. We consciously chose to avoid any benefits or penalties the Java compiler or run-time may provide to a familiar class such as String. We neither wanted to exploit any low-level implementation nor pay a cost for using a wrapper method to access specific characters in a String object.

```java
public int indexOf(final char[] target, final char[] pattern)
{
    final int targetLen = target.length;
    final int patternLen = pattern.length;
    final int lenDiff = targetLen - patternLen;
    int targetPos = -1;
    char scoutChar = '\u0000';
    int scoutPos = -1;
    int tmpPos = 0;

    // Loop through the target string.
    while (++targetPos <= lenDiff)
    {
        boolean scootUp = false;
        // Brute force move forward to find the next scout.
        int patternPos = -1;
        while (!scootUp && ++patternPos < patternLen &&
            (pattern[patternPos] == target[targetPos + patternPos]))
        {
            // If pattern position char is equal to scout char,
            // move target position forward.
            tmpPos = scoutPos - patternPos - 1;
            if (targetPos < tmpPos && pattern[patternPos] == scoutChar)
            {
                targetPos = tmpPos;
                scootUp = true;
            }
        }
        if (patternPos == patternLen)
            return targetPos;
        if (scootUp)
            continue;
```

```
            // If we're here, we found a scout. Let it march forward.
            tmpPos = lenDiff + patternPos; // Previous tmpPos is irrelevant.
            scoutChar = pattern[patternPos];
            scoutPos = targetPos + patternPos;
            while (++scoutPos <= tmpPos && scoutChar != target[scoutPos])
            { }
            if (scoutPos > tmpPos)
                return -1;
            // If we're here, the scout found a match.
            targetPos = scoutPos - patternPos - 1;
        }
        return -1;
}
```

# Appendix B: Real-life Testbed

We used a famous soliloquy from William Shakespeare's Hamlet, Act III, Scene I as our testbed. In our tests, we replaced the carriage returns with a single space so as to form a single large target string. The strings in bold font represent our test cases. Our test cases checked for the presence of those pattern strings verbatim including punctuation, spaces and mixed cases. The percentages on the right represent the depth at which the respective pattern string was found in the target string.

| | |
|---|---:|
| **To be, or not to be, that is the question**: Whether 'tis nobler in the mind to suffer | 0.00% |
| **The slings and arrows of outrageous fortune**, Or to take arms against a sea of troubles, | 5.83% |
| And by opposing end them? To die—to sleep, No more; and by a sleep to say we end | |
| The heart-ache, and the thousand natural shocks That flesh is heir to: 'tis a consummation | |
| Devoutly to be wish'd. To die, to sleep. **To sleep, perchance to dream—ay, there's the rub**, | 26.29%, 28.09% |
| For in that sleep of death what dreams may come, When we have **shuffled off this mortal coil**, | 33.92% |
| Must give us pause. There's the respect That makes calamity of so long life. | |
| For who would bear the whips and scorns of time, The oppressor's wrong, **the proud man's contumely**, | 46.10% |
| The pangs of despriz'd love, the law's delay, The insolence of office, and the spurns | |
| That patient merit of the unworthy takes, When he himself might his quietus make | |
| With a **bare bodkin**? Who would these fardels bear, To grunt and sweat under a weary life, | 59.17% |
| But that **the dread of something after death**, The undiscover'd country, from whose bourn | 66.35% |
| No traveller returns, puzzles the will, And makes us rather bear those ills we have | |
| Than fly to others that we know not of? Thus **conscience does make cowards of us all**, | 80.78% |
| And thus the native hue of resolution Is sicklied o'er with the pale cast of thought, | |
| And enterprises of great pith and moment, With this regard their currents turn awry | |
| And lose the name of action. Soft you now, The fair Ophelia! Nymph, in thy orisons | |
| **Be all my sins remember'd.** | 100% |

# Appendix C: Tabular Comparison of Various Algorithms

Although good performance is a desirable goal for any algorithm, we have shown that other factors may play an important role in the selection of an algorithm for general-purpose use. In the table below, we critique several algorithms. Some are our exemplars, which we have presented in detail. Others are originals or variants of the exemplars, which we have studied deeply. Yet others are algorithms we have studied but not analysed in depth. We present subjective findings in the table below. Our efforts here do not supplant other surveys. Rather, we present a quick guide that can direct further study.

| Algorithm | | |
|---|---|---|
| Criterion | Assessment | Comment |
| **Brute force** | | |
| Performance (wall-clock) | Moderate | The baseline |
| Memory Efficiency | High | No extra arrays |
| Locality | Moderate | Some spatial and temporal locality |
| Language Agnosticism | High | Works for any language as-is |
| Character Set Independence | High | Works for Unicode |
| Implementation Simplicity | High | Few lines of code |
| **Morris-Pratt [MP70], Knuth-Morris-Pratt [KMP77]** | | |
| Performance (wall-clock) | Moderate | Can be slightly better or worse depending on test |
| Memory Efficiency | Moderate | Requires prefix patterns to be stored |
| Locality | Low | Requires accessing the prefix array |
| Language Agnosticism | High | Works for any language as-is |
| Character Set Independence | High | Works for Unicode |
| Implementation Simplicity | Moderate | Preprocessing, intricate logic |
| **Karp-Rabin [KR87]** | | |
| Performance (wall-clock) | Low | Often worse than brute force |
| Memory Efficiency | High | No extra arrays |
| Locality | Moderate | Some spatial and temporal locality |
| Language Agnosticism | High | Works for any language as-is |
| Character Set Independence | Moderate | Hash function choice depends on char set |
| Implementation Simplicity | High | Few lines of code |
| **Rolling Sum, Rolling XOR** | | |
| Performance (wall-clock) | High | Often beats baseline |
| Memory Efficiency | High | No extra arrays |
| Locality | Moderate | Some spatial and temporal locality |
| Language Agnosticism | High | Works for any language as-is |
| Character Set Independence | High | Works for Unicode |
| Implementation Simplicity | High | Few lines of code |
| **Boyer-Moore [BM77], Turbo Boyer-Moore, Tuned Boyer-Moore, Horspool [Hor80], Smith [Smi91], Raita [Rai92], Sunday Quick [Sun90], Scout Sunday** | | |
| Performance (wall-clock) | High | Beats baseline for larger targets |
| Memory Efficiency | Moderate | Requires bad character array |
| Locality | Low | Requires accessing the prefix array |
| Language Agnosticism | High | Works for any language within defined character set |
| Character Set Independence | Low | Bad character array size depends on char set |
| Implementation Simplicity | Moderate | Preprocessing, intricate logic |
| **Zhu-Takaoka [ZT87], Berry-Ravindran [BR99], Reverse Colussi [Col94], Apostolico-Giancarlo [AG86], Skip, KMP Skip, Alpha Skip** | | |
| Performance (wall-clock) | High | Beats baseline for larger targets |
| Memory Efficiency | Moderate | Requires bad character array |
| Locality | Low | Requires accessing the prefix array |
| Language Agnosticism | High | Works for any language within defined character set |
| Character Set Independence | Low | Bad character array size depends on char set |
| Implementation Simplicity | Moderate | Preprocessing, intricate logic |
| **Sunday Maximal [Sun90], Sunday Optimal [Sun90]** | | |
| Performance (wall-clock) | High | Beats baseline for larger targets |
| Memory Efficiency | Moderate | Requires bad character array |
| Locality | Low | Requires accessing the prefix array |
| Language Agnosticism | Low | Requires character frequency in alphabet |
| Character Set Independence | Low | Bad character array size depends on char set |
| Implementation Simplicity | Moderate | Preprocessing, intricate logic |

| Scout, Scout Simple, Scout Twin, Scout Variant, Python Fast | | |
|---|---|---|
| Performance (wall-clock) | High | Often beats baseline |
| Memory Efficiency | High | No extra arrays |
| Locality | High | Accesses same or nearby characters |
| Language Agnosticism | High | Works for any language as-is |
| Character Set Independence | High | Works for Unicode |
| Implementation Simplicity | Moderate | Intricate logic |

## Appendix D: Performance Tables and Graphs

A key finding of our paper is that the Scout algorithm performs better than most alternative substring search algorithms. Our detailed empirical observations were made by implementing several algorithms in Java, taking care to neither advantage nor penalise any algorithm by using built-in classes and libraries. We also encoded our exemplar algorithms in Perl and C to verify if our finding holds. Broadly, it does. C and Perl are compiled or interpreted differently, so some of the absolute wall-clock times and crossover points change somewhat. Specifically, for the Hamlet testbed, the C and Perl implementations of Sunday Quick Search best the implementation of Scout.

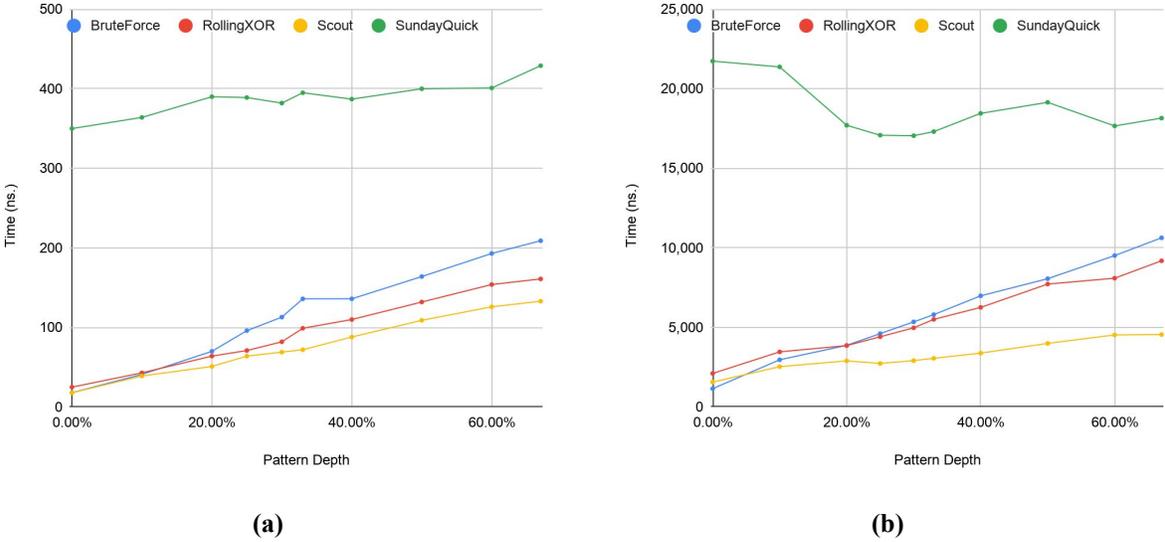

**(a)**      **(b)**

**Fig. 6:** Wall-clock times (ns.) for different pattern depths in a given target string length (100 characters). **(a)** C implementation. **(b)** Perl implementation.

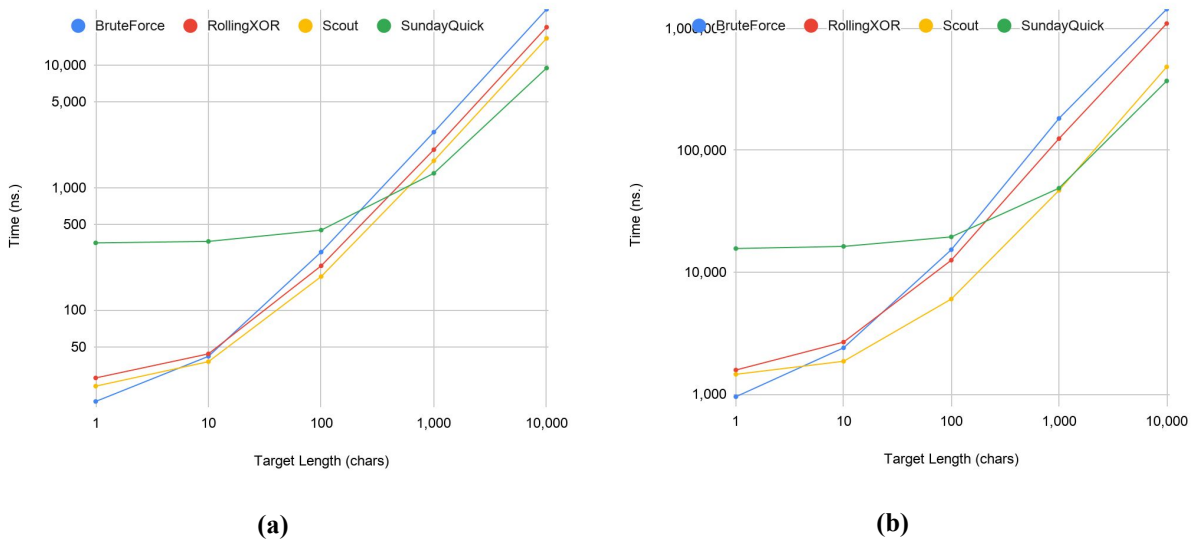

**Fig. 7:** Wall-clock times (ns.) for different target string lengths with a given pattern depth (100%). Log scales. **(a)** C implementation. **(b)** Perl implementation.

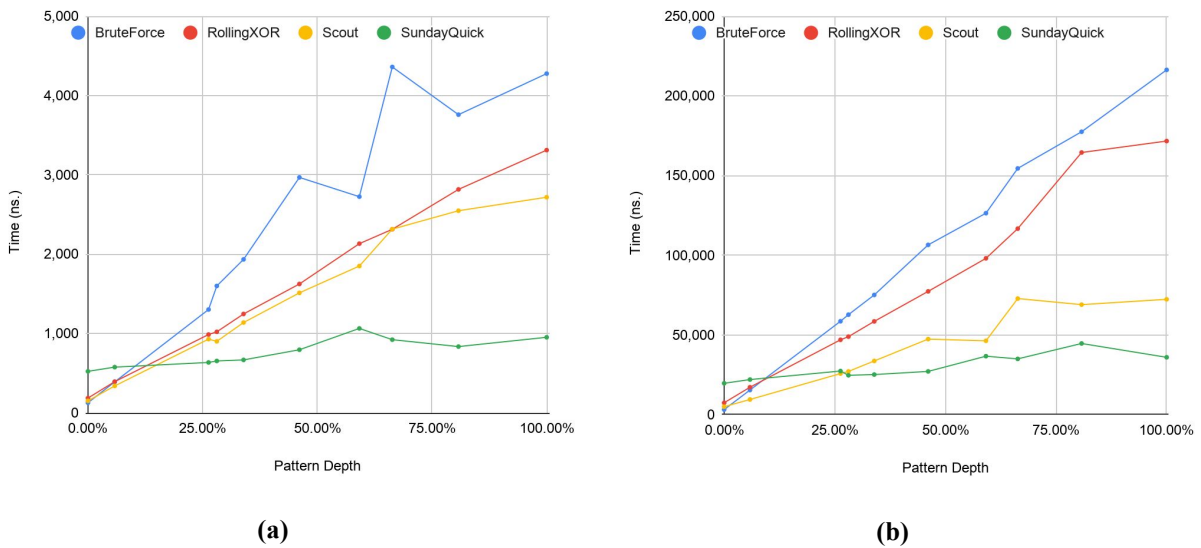

**Fig. 8:** Wall-clock times (ns.) for different pattern depths in Hamlet's soliloquy (1500 characters). **(a)** C implementation. **(b)** Perl implementation.

Through most of this paper, we have chosen four exemplar algorithms - Brute force, Rolling XOR, Sunday Quick Search and Scout - in order to compare performance numbers. We chose these four so as to compare best-in-class algorithms and present the results in a readable manner. We have maintained that our unpublished results support our broad conclusions. In the tables below, we show character comparisons, memory lookups and wall-clock times for the Hamlet testbed for Java implementations of several algorithms and variants. The character comparisons and memory lookups have been averaged out for our one million runs, but the wall-clock time is simply the sum over all million runs. These differences merely aid presentation, but do not affect the overall observations and conclusions. The exemplar columns have been highlighted for ease of perusal.

**Table. 1:** Character comparisons for different pattern depths in Hamlet's soliloquy (1500 characters).

| Pattern Depth | BoyerMoore | Brute Force | Horspool | Karp Rabin | Knuth MorrisPratt | Rolling Sum | Rolling XOR | Scout | Scout Simple | Scout Simple Twin | Scout Sunday | Scout Twin | Scout Var | Sunday Quick |
|---|---|---|---|---|---|---|---|---|---|---|---|---|---|---|
| 0.00% | 41 | 41 | 41 | 41 | 121 | 41 | 41 | 41 | 41 | 52 | 41 | 462 | 41 | 41 |
| 5.83% | 49 | 129 | 49 | 43 | 297 | 43 | 45 | 131 | 137 | 56 | 80 | 643 | 132 | 53 |
| 26.29% | 56 | 427 | 56 | 35 | 856 | 29 | 31 | 439 | 425 | 453 | 298 | 695 | 445 | 58 |
| 28.09% | 69 | 460 | 68 | 22 | 887 | 23 | 22 | 453 | 442 | 461 | 329 | 576 | 459 | 53 |
| 33.92% | 61 | 560 | 61 | 31 | 1,084 | 29 | 34 | 565 | 542 | 303 | 359 | 823 | 577 | 66 |
| 46.10% | 89 | 780 | 90 | 34 | 1,434 | 26 | 27 | 777 | 714 | 739 | 431 | 986 | 794 | 71 |
| 59.17% | 128 | 901 | 130 | 23 | 1,794 | 11 | 24 | 914 | 928 | 939 | 826 | 965 | 921 | 126 |
| 66.35% | 100 | 1,119 | 100 | 46 | 2,050 | 38 | 39 | 1,122 | 1,028 | 1,063 | 742 | 1,375 | 1,150 | 111 |
| 80.78% | 101 | 1,244 | 110 | 44 | 2,477 | 39 | 41 | 1,262 | 1,256 | 1,293 | 984 | 1,677 | 1,268 | 103 |
| 100.0% | 111 | 1,501 | 111 | 38 | 3,024 | 33 | 37 | 1,504 | 1,506 | 1,532 | 1,463 | 1,732 | 1,505 | 114 |

**Table. 2:** Memory lookups for different pattern depths in Hamlet's soliloquy (1500 characters).

| Pattern Depth | BoyerMoore | Brute Force | Horspool | Karp Rabin | Knuth MorrisPratt | Rolling Sum | Rolling XOR | Scout | Scout Simple | Scout Simple Twin | Scout Sunday | Scout Twin | Scout Var | Sunday Quick |
|---|---|---|---|---|---|---|---|---|---|---|---|---|---|---|
| 0.00% | 419 | 82 | 417 | 164 | 283 | 164 | 164 | 82 | 82 | 94 | 419 | 991 | 82 | 419 |
| 5.83% | 451 | 258 | 449 | 342 | 638 | 342 | 346 | 179 | 197 | 103 | 473 | 1,274 | 177 | 463 |
| 26.29% | 479 | 854 | 477 | 900 | 1,746 | 888 | 892 | 503 | 470 | 492 | 688 | 1,067 | 491 | 485 |
| 28.09% | 529 | 920 | 519 | 914 | 1,817 | 916 | 914 | 501 | 474 | 488 | 683 | 784 | 490 | 467 |
| 33.92% | 499 | 1,120 | 497 | 1,118 | 2,227 | 1,114 | 1,124 | 654 | 599 | 344 | 779 | 1,236 | 631 | 515 |
| 46.10% | 609 | 1,560 | 611 | 1,478 | 2,940 | 1,462 | 1,464 | 905 | 756 | 774 | 874 | 1,389 | 872 | 539 |
| 59.17% | 753 | 1,802 | 753 | 1,830 | 3,609 | 1,806 | 1,832 | 962 | 1,011 | 987 | 1,163 | 1,091 | 947 | 733 |
| 66.35% | 655 | 2,238 | 653 | 2,106 | 4,204 | 2,090 | 2,092 | 1,321 | 1,104 | 1,120 | 1,279 | 1,935 | 1,268 | 691 |
| 80.78% | 653 | 2,488 | 687 | 2,526 | 5,010 | 2,516 | 2,520 | 1,366 | 1,368 | 1,369 | 1,417 | 2,281 | 1,338 | 661 |
| 100.0% | 699 | 3,002 | 697 | 3,076 | 6,075 | 3,066 | 3,074 | 1,537 | 1,544 | 1,565 | 1,809 | 2,033 | 1,534 | 711 |

**Table. 3:** Wall-clock times (ns.) for different pattern depths in Hamlet's soliloquy (1500 characters).

| Pattern Depth | BoyerMoore | Brute Force | Horspool | Karp Rabin | Knuth MorrisPratt | Rolling Sum | Rolling XOR | Scout | Scout Simple | Scout Simple Twin | Scout Sunday | Scout Twin | Scout Var | Sunday Quick |
|---|---|---|---|---|---|---|---|---|---|---|---|---|---|---|
| 0.00% | 161 | 54 | 168 | 275 | 187 | 51 | 56 | 63 | 35 | 41 | 173 | 437 | 67 | 154 |
| 5.83% | 168 | 272 | 171 | 602 | 452 | 170 | 194 | 107 | 114 | 50 | 226 | 600 | 109 | 234 |
| 26.29% | 262 | 1,063 | 264 | 1,582 | 1,306 | 554 | 708 | 268 | 262 | 286 | 357 | 677 | 265 | 279 |
| 28.09% | 331 | 1,128 | 314 | 1,730 | 1,337 | 609 | 759 | 255 | 265 | 296 | 332 | 537 | 255 | 280 |
| 33.92% | 265 | 1,373 | 269 | 2,010 | 1,636 | 702 | 881 | 383 | 338 | 209 | 446 | 829 | 350 | 282 |
| 46.10% | 425 | 1,912 | 418 | 3,134 | 2,173 | 974 | 1,149 | 538 | 427 | 472 | 534 | 1,004 | 547 | 366 |
| 59.17% | 588 | 2,427 | 591 | 3,508 | 2,780 | 1,178 | 1,600 | 447 | 613 | 667 | 530 | 832 | 433 | 604 |
| 66.35% | 447 | 2,730 | 491 | 4,059 | 3,234 | 1,365 | 1,690 | 804 | 641 | 705 | 759 | 1,612 | 775 | 507 |
| 80.78% | 406 | 3,473 | 465 | 4,903 | 3,761 | 1,650 | 2,070 | 657 | 828 | 913 | 668 | 1,543 | 639 | 449 |
| 100.0% | 515 | 3,949 | 522 | 6,133 | 4,432 | 2,076 | 2,593 | 612 | 1,006 | 1,080 | 713 | 1,451 | 619 | 557 |

We choose to defer publication of these results so that we may implement all of the forty or so available algorithms and perform a thorough comparison.